\documentclass[runningheads]{llncs}
\usepackage{graphicx}
\usepackage{amsmath,amssymb,amsfonts}
\usepackage{cite}
\begin{document}
\title{Mesarovician Abstract Learning Systems}
%
%
\author{Tyler Cody}
\authorrunning{T. Cody}
%
\institute{National Security Institute \\ Virginia Tech, Arlington Virginia, USA \\
\email{tcody@vt.edu}}
\maketitle              
\begin{abstract}

The solution methods used to realize artificial general intelligence (AGI) may not contain the formalism needed to adequately model and characterize AGI. In particular, current approaches to learning hold notions of problem domain and problem task as fundamental precepts, but it is hardly apparent that an AGI encountered in the wild will be discernable into a set of domain-task pairings. Nor is it apparent that the outcomes of AGI in a system can be well expressed in terms of domain and task, or as consequences thereof. Thus, there is both a practical and theoretical use for meta-theories of learning which do not express themselves explicitly in terms of solution methods. General systems theory offers such a meta-theory. Herein, Mesarovician abstract systems theory is used as a super-structure for learning. Abstract learning systems are formulated. Subsequent elaboration stratifies the assumptions of learning systems into a hierarchy and considers the hierarchy such stratification projects onto learning theory. The presented Mesarovician abstract learning systems theory calls back to the founding motivations of artificial intelligence research by focusing on the thinking participants directly, in this case, learning systems, in contrast to the contemporary focus on the problems thinking participants solve.

\end{abstract}

\section{Notation}

The Cartesian product is denoted $\times$. Given system $S \subset \times \{V_i\}$ for i = 0, ..., I, $\overline{S}$ denotes the component sets of $S$, i.e., $\overline{S} = \{V_0, ..., V_I\}$.

\section{Introduction}

Artificial intelligence (AI) was initiated as a field to study the realization of thinking in computers \cite{mccarthy2006proposal}. Over the years, however, AI’s focus has drifted away from thinking in general towards problem solving in particular. In learning, this is epitomized by the near-universal precepts of problem domain and problem task \cite{thorisson2016artificial}. Typically, the domain and task are formalized as $\mathcal{D} = \{\mathcal{X}, P(X)\}$ and $\mathcal{T} = \{\mathcal{Y}, P(Y|X)\}$ where $\mathcal{X}$ and $\mathcal{Y}$ are the inputs and outputs of a function that an AI is approximating. This view, taken to the extreme, posits intelligence as a problem-solving phenomenon to be measured by integrating an error function over a complexity-weighted set of domain-task pairings \cite{chollet2019measure}.

Artificial general intelligence (AGI) is more than problem-solving, however. And, in engineering AGI, the problems AGI solves are merely part of broader systems concerns. Viewing learning through the lens of problems makes systems concerns at least secondary. And, moreover, relying on domain and task as precepts greatly limits the extent to which formalism can be carried through into general elaborations. As AI is largely a mathematical construct, the use of metaphors and analogies in the stead of axoims and first principles is unnecessary for its basic systems characterization---and a basic systems characterization may be all one can hope to achieve as the influence and outcomes of AGI will likely not be readily discernable, let alone discernable into domain-task pairings.

While perhaps parsimonious for describing solution methods, notions of domain and task lack the formalism needed for extensive elaboration at a general level and are insufficient for characterizing AGI as a system or the roles AGI plays in systems. Mesarovician abstract systems theory (AST) can be used to address these short-comings by treating learning as a system, as opposed to as a problem-solving procedure. This manuscript contributes a Mesarovician abstract learning systems theory (ALST) that builds upon previous work in transfer learning \cite{cody2021systems} with notions of hierarchy. Namely, abstract learning systems are stratified in order of the generality of their assumptions and such stratification is projected onto learning theory.

The manuscript is structured as follows. First, relevance to AGI is motivated and preliminaries on AST and ALST are given. Then, learning systems are stratified, that stratification is projected onto learning theory, and, before concluding, remarks are made on scope and practical use.

\section{Motivation}

There are a number of fundamental challenges to modeling AGI.
\begin{itemize}
    \item Perhaps AGI will be realized by a well-formulated problem domain and task coupled with an explicit solution method \cite{thorisson2016artificial}. But, even if this is the case, requisite variety \cite{ashby1961introduction} and chaos \cite{lorenz1963deterministic} suggest that any solution method capable of realizing AGI will require an abstraction mechanism burdened by nearly unbounded variety and irreducible complexity. So, one will not be able to reliably look at a solution method and foretell its outcomes or look at outcomes and discern the solution method.
    \item It may also be that AGI is an emergent phenomena among a system of systems \cite{goertzel2017formal}, incompressible into individual solution methods, let alone domain-task pairings. In such a case, AGI phenomena exist at a higher level of abstraction than the individual solution methods themselves.
    \item AGI is expected to influence and to be influenced by the system within which it operates. This coupling suggests that even if AGI can be relegated to a sub-system at conception, the borders between the ``intelligent'' sub-system and those under its influence face dissolution as the AGI and its context intertwine. Thus, it may be that, after a period of integration, an AGI's solution method is not representative of the form the AGI comes to take.
\end{itemize}

\begin{figure}[t]
    \centering
    \includegraphics[width=0.8\textwidth]{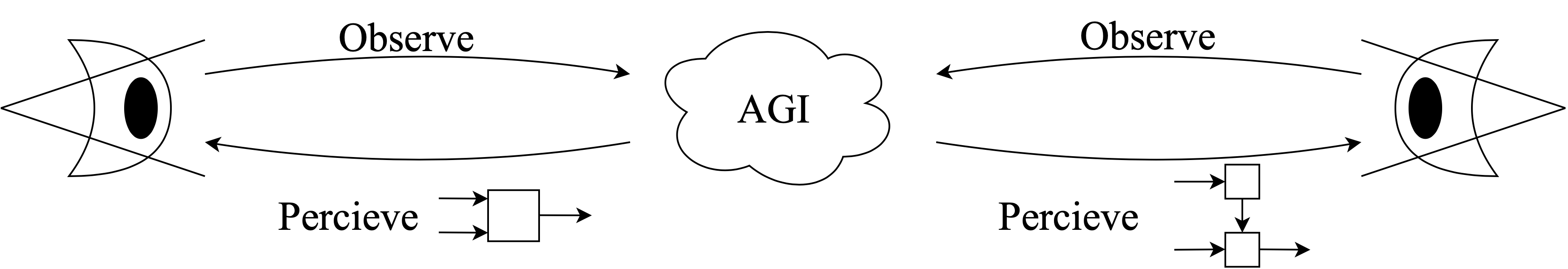}
    \caption{Observation is an active process influenced by the values of the observer \cite{ahl1996hierarchy}.}
    \label{fig:observation}
\end{figure}

And so, for these reasons, it seems natural to study learning in AGI in terms of general systems phenomena. The high level of abstraction allows for a stratification in the specification of assumptions when modeling learning, thereby allowing for significant, formal elaboration without explicit reference to solution methods. This stratification allows for addressing uncertainty in modeling by choice of perspective view. It supports modeling the AGI phenomena one observes, i.e., the phenomena one's values lead them to perceive \cite{ahl1996hierarchy}, as depicted in Figure \ref{fig:observation}. Thus, the presented theory provides a far-reaching, formal framework for learning that observers and engineers of AGI can use to scope their field of view.

\section{Preliminaries}

\subsection{AST}

General systems theory is concerned with the study of phenomena that apply to systems in general \cite{bertalanffy1969general}. AST is a mathematical general systems theory born out of systems engineering \cite{mesarovic1989abstract}. However, it quickly found cross-disciplinary application, notably, in fields of information, computation, and cybernetics \cite{mesarovic1975general, rine1971categorical}. 

\begin{figure}[htb]
    \centering
    \includegraphics[width=0.5\textwidth]{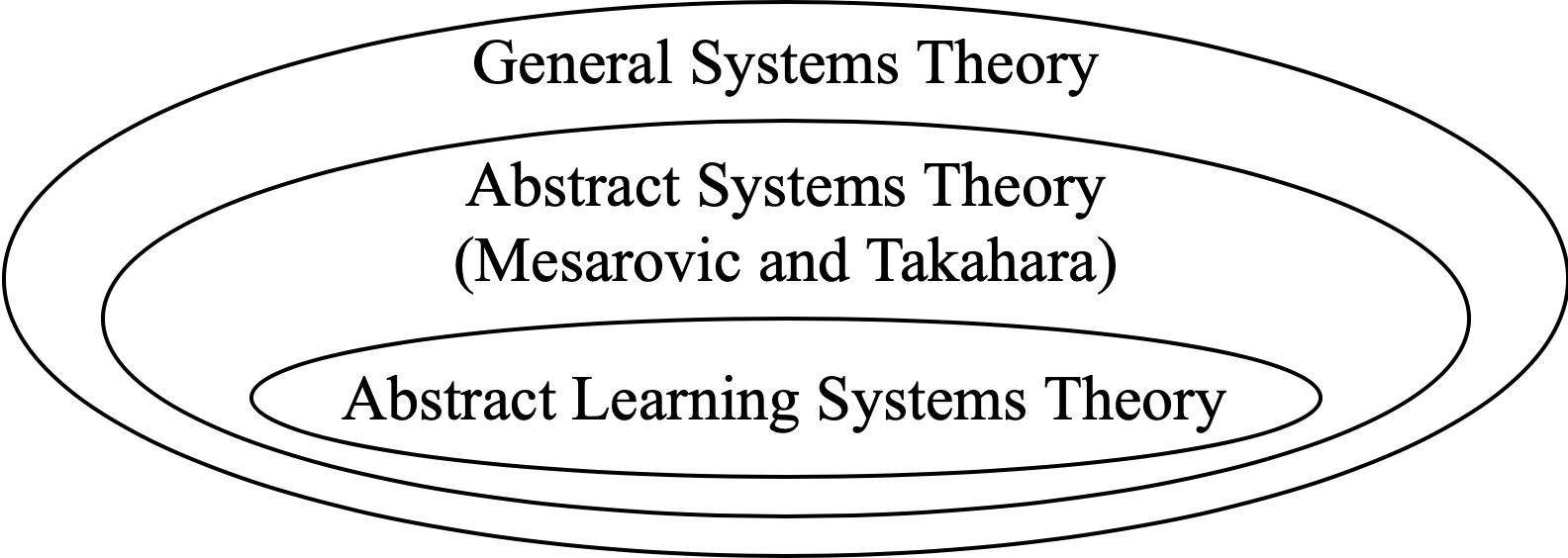}
    \caption{Venn diagram contextualizing the presented theory.}
    \label{fig:venn}
\end{figure}

In the realm of general systems theory, it holds the formal-minimalist worldview that systems are a relation on sets \cite{dori2019system}. Mesarovic and Takahara posit AST as an attempt to formalize block-diagrams without a loss of generality---that is, as a formal, intermediate step between verbal descriptions and detailed mathematical models \cite{mesarovic1975general}. Abstract learning systems theory is depicted in context in Figure \ref{fig:venn}.

In AST, a system is defined as a relation $S \subset \times \{V_i\}$ where $V_i$ for $i \in I$ are termed component sets. Theory is developed by adding structure to the component sets, their elements, and the relation among them. Two fundamental systems in AST are input-output systems and goal-seeking systems. An \emph{input-output} or \emph{elementary} system is a relation $S \subset \mathcal{X} \times \mathcal{Y}$ where $\mathcal{X} \cap \mathcal{Y} = \emptyset$ and $\mathcal{X} \cup \mathcal{Y} = \overline{S}$. When, $S:\mathcal{X} \to \mathcal{Y}$, $S$ is termed a \emph{functional} system.

A \emph{goal-seeking} system is an input-output system $S \subset \mathcal{X} \times \mathcal{Y}$ with internal feedback. The internal feedback is specified by a set of consistency relations $G: \mathcal{X} \times \mathcal{Y} \times \Theta \to V$ and $E: \mathcal{X} \times \mathcal{Y} \times V \to \Theta$. $G$ is termed the goal relation and is responsible for assigning values $v \in V$ to input-output pairs $(x, y) \in \mathcal{X} \times \mathcal{Y}$. $E$ is termed the search or seeking relation and is responsible for selecting the internal control parameter $\theta \in \Theta$. Importantly, $G$ and $E$ cannot be composed to form $S$---in other words, goal-seeking systems are input-output systems at their highest level of abstraction, but cannot be specified as a composition of input-output systems.

\subsection{ALST}

Recent work has extended AST into transfer learning \cite{cody2021systems}. There, transfer learning was modeled as a relation on learning systems, and notions of transferability, transfer distance, and transfer roughness were defined in systems theoretic terms. In contrast, this manuscript establishes ALST as a general systems theory concerned with learning broadly.

Learning systems are formulated as a cascade connection of a goal-seeking system and an input-output system. \emph{Learning} systems are defined as follows \cite{cody2021systems}.
\begin{definition}{\emph{Learning System}.} \\
    A learning system $S$ is a relation
    $$S \subset \times \{ A, D, \Theta, G, E, H, \mathcal{X}, \mathcal{Y} \}$$
    such that
    \begin{gather*}
        D \subset \mathcal{X} \times \mathcal{Y}, A:D \to \Theta, H:\Theta \times \mathcal{X} \to \mathcal{Y} \\
        (d, x, y) \in \mathcal{P}(S) \leftrightarrow (\exists \theta) [(\theta, x, y) \in H \wedge (d, \theta) \in A] \\
        G: D \times \Theta \to V, E: V \times D \to \Theta \\
        (d, G(\theta, d), \theta) \in E \leftrightarrow (d, \theta) \in A
    \end{gather*}
    where
    $$x \in \mathcal{X}, y \in \mathcal{Y}, d \in D, \theta \in \Theta.$$
    The algorithm $A$, data $D$, parameters $\Theta$, consistency relations $G$ and $E$, hypotheses $H$, input $\mathcal{X}$, and output $\mathcal{Y}$ are the component sets of $S$, and learning is specified in the relation among them.
    \label{def:gsls}
\end{definition} 

Learning systems can be decomposed into two systems $S_I$ and $S_F$. The inductive system $S_I \subset \times \{ A, D, \Theta \}$ is responsible for inducing hypotheses from data. The functional system $S_F \subset \times \{ \Theta, H, \mathcal{X}, \mathcal{Y} \}$ is the induced hypothesis. $S_I$ and $S_F$ are coupled by the parameter $\Theta$. Learning is hardly a purely input-output process, however, and, to address this, the goal-seeking nature of $S_I$, and, more particularly, of $A$ is specified. $A$ is goal-seeking in that it makes use of a goal relation $G: D \times \Theta \to V$ that assigns a value $v \in V$ to data-parameter pairs, and a seeking relation $E: V \times D \to \Theta$ that assigns a parameter $\theta \in \Theta$ to data-value pairs. Again, note, these consistency relations $G$ and $E$ specify $A$, but are not a decomposition of $A$. Also note that $D$ is specified as a subset of $\mathcal{X} \times \mathcal{Y}$ following convention, not necessity.

\section{Stratification}

In the following, the hierarchy of assumptions in this formulation of learning systems as well as the hierarchy it projects onto learning theory are investigated.

\subsection{Levels in Abstract Learning Systems Theory}

\begin{figure}[t]
    \centering
    \includegraphics[width=0.85\textwidth]{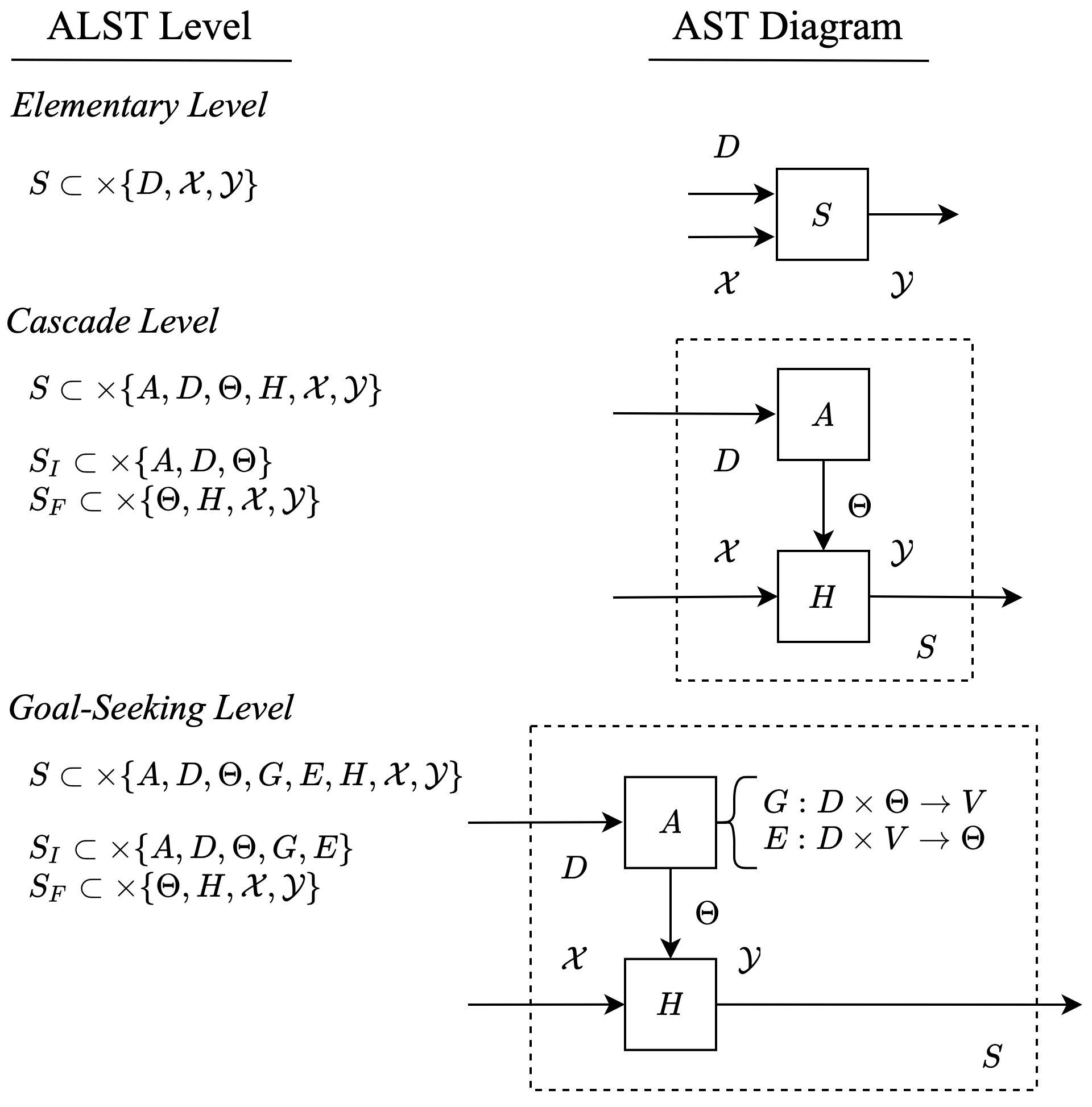}
    \caption{The three stratified levels in terms of their component sets and block-diagrams.}
    \label{fig:ast-strat}
\end{figure}

Each component set of a learning system can be modeled with considerable depth. But, when taking a top-down view, there are three key levels of abstraction implicit in Definition \ref{def:gsls} and depicted in Figure \ref{fig:ast-strat}.

\subsubsection{Elementary Level.}

The elementary level treats learning as an input-output system $S \subset \times \{D, \mathcal{X}, \mathcal{Y}\}$. Specifically, $S:D \times \mathcal{X} \to \mathcal{Y}$. This level is already sufficient to characterize a learning system in terms the fundamental properties that AST is built upon. For example, the \emph{stability} of $S(D):\mathcal{X} \to \mathcal{Y}$, whether $\mathcal{X}$ or $D$ is \emph{anticipatory} of $\mathcal{Y}$, or whether $S(D):\mathcal{X} \to \mathcal{Y}$ is \emph{controllable} by $D$. Also, this level admits consideration of the \emph{composition} or \emph{interaction} of $S$ with other systems in its context. This level, however, is restricted to little more than the analysis of correlation between inputs and outputs.

\subsubsection{Cascade Level.}

The cascade level treats learning as a cascade connection of input-output systems, particularly, a cascade connection of an inductive system $S_I$ and the hypotheses it induces $S_F$, i.e., $S \subset \overline{S_I} \times \overline{S_F}$ where $S_I \subset \times \{A, D, \Theta\}$ and $S_F \subset \times \{\Theta, H, \mathcal{X}, \mathcal{Y}\}$. More specifically, $S \subset \times \{A, D, \Theta, H, \mathcal{X}, \mathcal{Y}\}$ where $A:D \to \Theta$ and $H: \Theta \times \mathcal{X} \to \mathcal{Y}$. Learning systems are still input-output systems, as at the elementary level, but now the model of learning distinguishes the inductive part as cascading into the functional part.

\subsubsection{Goal-Seeking Level.}

The goal-seeking level treats learning as a cascade connection of a goal-seeking system and an input-output system. The inductive system is extended to specify its goal-seeking nature, i.e., $S_I \subset \times \{A, D, \Theta, G, E\}$ where $G:D \times \Theta \to V$ and $E:D \times V \to \Theta$. The explicit consideration of the goal-seeking nature of learning distinguishes this level from higher levels as it specifies that $S_I$ is not decomposable into input-output systems. As such, this level acknowledges that the traditional engineering practice of engineering by aggregation, of following the mantra, ``If the parts work, the whole will work'', will not necessarily work for engineering learning.

\subsection{Remarks on Levels}

The goal-seeking level may seem in conflict with the higher levels of abstraction. It is not. At the elementary and cascade levels, learning is appropriately addressable as an input-output system. Such a view, of course, treats goal-seeking as a black-box. Unpacking the black-box, goal-seeking nature of learning can be done at a lower level of abstraction, but not without sacrificing the simplicity of learning as an elementary system or as a cascade of elementary systems.

\subsection{Projection of Levels Onto Learning Theory}

\begin{figure}[t]
    \centering
    \includegraphics[width=0.85\textwidth]{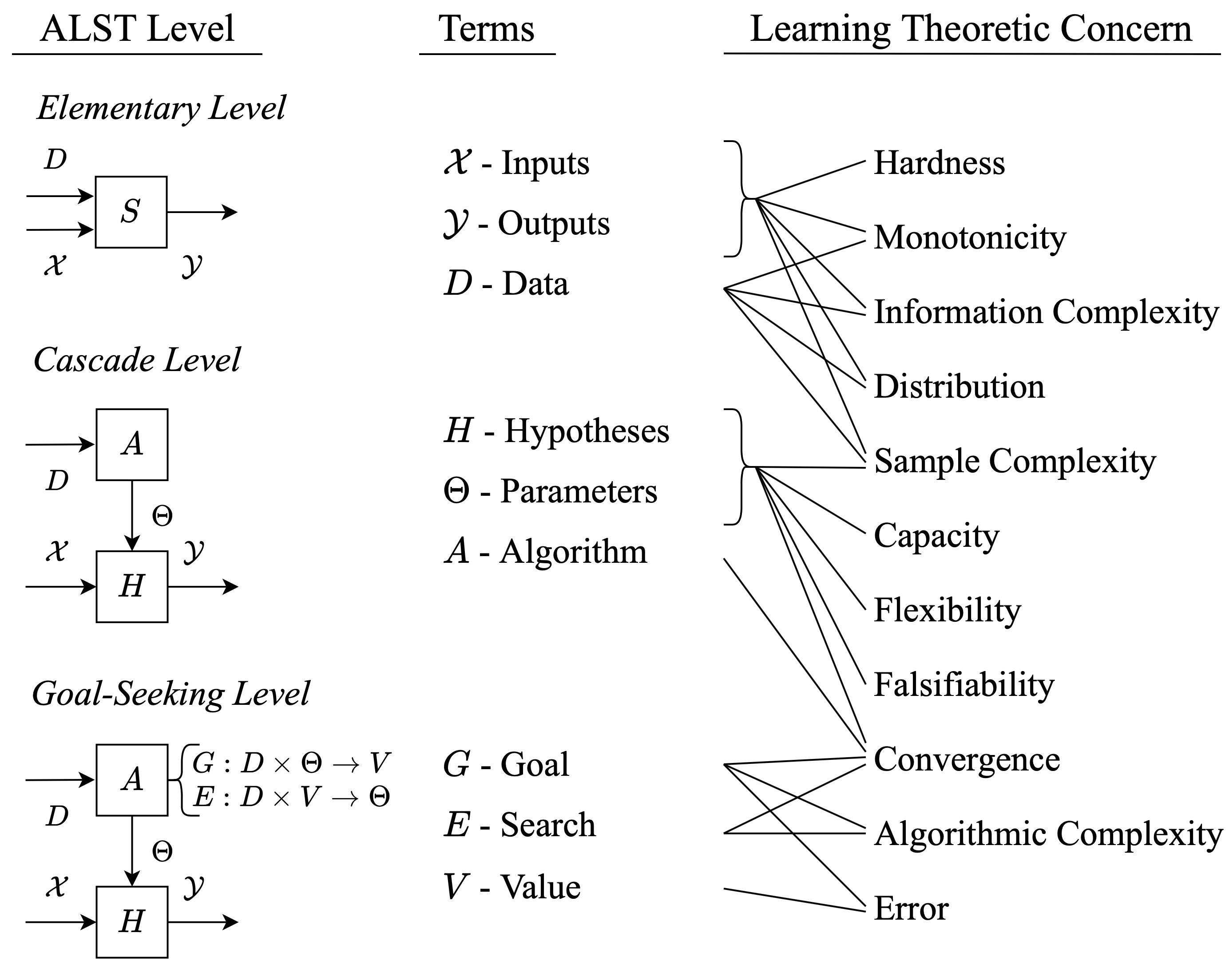}
    \caption{The three stratified levels in terms of their block-diagrams, the additional component sets they consider, and the relationship between those component sets and learning theoretic concerns.}
    \label{fig:stratification}
\end{figure}

This stratification of learning systems projects a stratification onto the many concerns of learning theory. We demonstrate this using the 11 concerns listed in the right column of Figure \ref{fig:stratification}. In Figure \ref{fig:stratification}, the component sets of each level are associated with each of these learning theoretic concerns. 

\subsubsection{Elementary Level.}

The elementary level, concerning $D, \mathcal{X}$ and $\mathcal{Y}$, allows for the most general phenomena to be considered. Learning problems have a \emph{hardness} associated with what the learning system is tasked to do given $\mathcal{X}$. If $\mathcal{X}$ are paths in a graph and $\mathcal{Y}$ is the longest path in the set of paths $\mathcal{X}$, then the learning problem is NP-complete. Given $D$ in addition to $\mathcal{X}$ and $\mathcal{Y}$, the basic properties of \emph{monotonicity} can be investigated and
\emph{information complexity} can be estimated, along with other basic considerations of \emph{distribution}. Learning theory at the elementary level gives a sense of how hard a learning problem is and how much there is to learn, but no particulars regarding the inner-workings of the learning system itself.

\subsubsection{Cascade Level.}

The cascade level introduces hypotheses $H$ and its parameterization $\Theta$, which gives a sense of the terms in which a learning system interprets the world, or at least gives a sense of the form of its explanations of worldly phenomena. Given $H$ and $\Theta$, notions of \emph{falsifiability}, \emph{flexibility}, and \emph{capacity} can be considered. Falsifiability refers to whether the hypotheses are suited for scientific induction. Flexibility refers to the rate at which the relation between $\mathcal{X}$ and $\mathcal{Y}$ specified by $H(\Theta)$ can change when $\Theta$ is varied. And capacity refers to the variety with which a set of hypotheses can partition $\mathcal{X}$, i.e., capacity concerns how many different labelings of $\mathcal{X}$ by $\mathcal{Y}$ are possible using $H(\Theta)$. Additionally, at the cascade level, \emph{sample complexity} can be modeled as a distribution-free property of hypotheses $H(\Theta)$ or, using distribution information from the elementary level, can be modeled as a property of hypotheses $H(\Theta)$ and the distribution over which they are induced. Capacity and sample complexity see particularly widespread use in learning theory. Some of the most important theorems concerning a learning systems ability to adapt to change are given in terms of sample complexity, capacity, and distributional divergence \cite{ben2010theory}. 

\subsubsection{Goal-Seeking Level.}

Although the cascade level introduces the algorithm $A$, $A$ is left as black-box. The goal-seeking level, however, allows for a detailed characterization of a learning systems goal-seeking nature. This includes some of the most common interests of learning theory, e.g., non-asymptotic \emph{convergence}, i.e., whether a system can learn a function approximation in the short or medium term, and error, i.e., the similarity between the approximated function and the induced hypothesis. Convergence is a statement made using $A, G,$ and $E$, as well as the hypotheses $H$ and how they are parameterized by $\Theta$. Error concerns the set of values $V$ and the goal relation $G$ that relates data $D$ and parameters $\Theta$ to those values. And, knowing how the search problem is formulated via $G$ and $E$, statements can be made on the \emph{algorithmic complexity} of $A$.

\subsection{Remarks on Projection of Levels}

While one can model learning directly using any individual or combination of the stratified component sets, there is always a systems theoretic structure implicit. By specifying that the hypotheses and error take a certain form, for example, one is also specifying, implicitly, something regarding $\mathcal{X}, \mathcal{Y}, \Theta, A,$ and $G$.

Capacity as a cascade level notion is worth elaborating with a particular example. Bottou and Vapnik define a notion of local learning \emph{algorithms} using capacity \cite{bottou1992local}. There, they define local algorithms as those that, ``...attempt to locally adjust the capacity of the training system to the properties of the training set in each area of the input space.'' In ALST terms, local learning \emph{systems} are those that, informed by $D$, adjust the capacity of $H(\Theta)$ during training in accordance with $\mathcal{X}$. Thus, local learning is a cascade level notion.

\section{Scope}

\begin{figure}[t]
    \centering
    \includegraphics[width=0.85\textwidth]{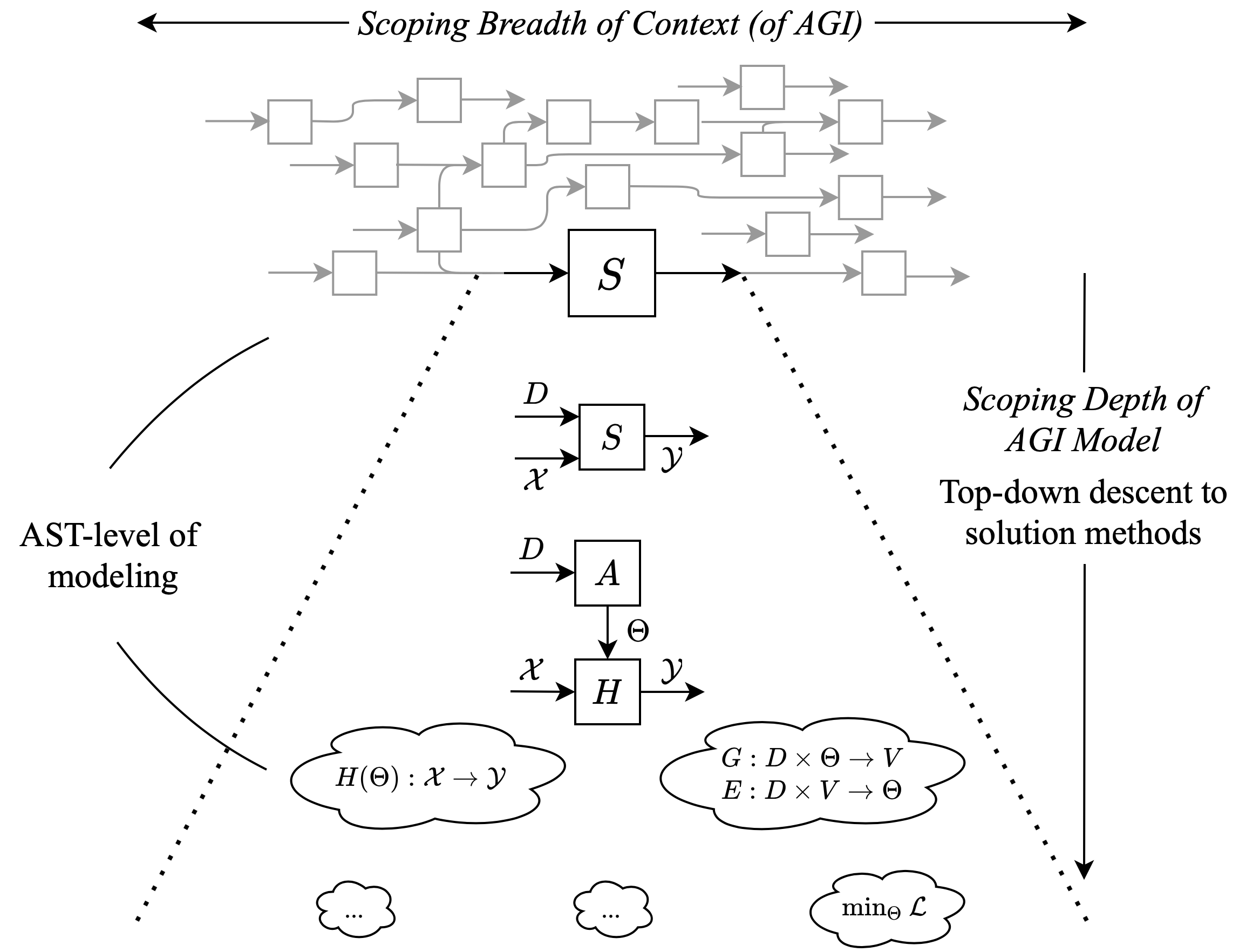}
    \caption{A depiction of the horizontal and vertical scoping of models of AGI afforded by ALST.}
    \label{fig:scope}
\end{figure}

ALST provides a mechanism for scoping the field of view with which AGI is modeled. The scene of grey input-output systems horizontally surrounding the black input-output system $S$ at the top of Figure \ref{fig:scope} depicts a learning system being contextualized by other system-level, input-output phenomena. The ability to scope a model of AGI outward into the AGI's context is inherited from AST. The vertical descent in Figure \ref{fig:scope} from the input-output system $S$ to the elementary level, then the cascade level, and so on, depicts the top-down scoping of depth in a model of AGI. This top-down descent from the general systems level towards solution methods is provided by ALST.

By stratifying learning systems with respect to the generality of their mathematical structure, learning systems can be specified at varying levels of abstraction. This can serve as a useful mechanism for engineering practice. A formal understanding of hierarchy allows for an ordering of design decisions, for the structuring of operational and mission performance models, and for modeling a learning system with various degrees of precision and uncertainty.

In engineering AGI, these are important capabilities. AGI solutions are overwhelmingly bottom-up, but the outcomes we associate with the success of AGI will occur at higher-levels of abstraction. And so, from the perspective of an engineer trying to build or use AGI towards satisfying the needs and goals of a stakeholder, the presented framework allows for the outcomes of AGI, the general characteristics of AGI, and the needs and goals of its stakeholders to be modeled in a common language and at a common level of abstraction.

\section{Conclusion}

The general systems theory approach to learning presented herein calls back to the principal concerns of AI’s founding, with a perspective view on learning in favor of the thinking participants themselves, not the problems they solve. It does this by shifting from a view of learning as problem solving to a view of learning as a system. And, in doing so, it provides a means of mathematically characterizing learning in AGI in terms of its general phenomena---without explicit reference to solution methods.

In addition to stratification, AST offers a number of other promising uses for AI. AST is largely a theory of category, and thereby provides a non-conventional means of applying category to AI. Also, AST provides the foundational mathematics for efforts in model-based systems engineering and digital engineering. AST may be a means of connecting AI to large-scale, formal models of systems developed by engineers. Lastly, while 3 key, hierarchical levels were emphasized herein, there are also a variety of heterarchical relationships to explore. A top-down understanding of these varied relationships may serve well as a  latticework about which to structure best practices for the engineering of AI.

\bibliographystyle{splncs04}
\bibliography{ref}

\end{document}